\documentclass[pdflatex,sn-mathphys-num]{sn-jnl}% Math and Physical Sciences Numbered Reference Style 
\usepackage{graphicx}%
\usepackage{multirow}%
\usepackage{amsmath,amssymb,amsfonts}%
\usepackage{amsthm}%
\usepackage{mathrsfs}%
\usepackage[title]{appendix}%
\usepackage{xcolor}%
\usepackage{textcomp}%
\usepackage{manyfoot}%
\usepackage{booktabs}%
\usepackage{algorithm}%
\usepackage{algorithmicx}%
\usepackage{algpseudocode}%
\usepackage{listings}%
\usepackage{multibib}
\newcites{Methods}{References}
\usepackage{amsfonts, amsmath, amssymb}
\usepackage{float}
\usepackage{dirtytalk}
\usepackage{subcaption}
\usepackage{graphicx}
\usepackage{booktabs}
\usepackage[svgnames,table, dvipsnames]{xcolor}
\usepackage[tableposition=above]{caption}
\usepackage{multirow}
\usepackage{colortbl}
\usepackage{adjustbox}
\usepackage{pifont}
\usepackage{footnote}
\usepackage{placeins}
\usepackage{hyperref}
\usepackage{rotating}

\hypersetup{colorlinks=false, linkbordercolor={white}, citebordercolor={white}, hidelinks}
\makeatletter
\newcommand{\listofsupptables}{%
  \section*{List of Tables} 
  \@starttoc{los} % This tells LaTeX to read from the .los file
}

\newcommand{\beginsupplement}{%
  \setcounter{table}{0} % Reset table counter
  \renewcommand{\thetable}{\arabic{table}} % Change numbering to S1, S2...
  \renewcommand{\ext@table}{los} % Crucial: redirect captions to .los file
}
\makeatother

\begin{document}

\title[Memorisation bias in medical AI]{Memorisation bias in medical AI}


\author*[1]{\fnm{Moritz A.} \sur{Knolle}}\email{moritz.knolle@tum.de}

\author[1,2]{\fnm{Martin J.} \sur{Menten}}

\author[1,2]{\fnm{Laurin} \sur{Lux}}

\author[]{\fnm{Mélanie} \sur{Roschewitz}}

\author[3]{\fnm{Emma A.M.} \sur{Stanley}}

\author[4]{\fnm{Georgios} \sur{Kaissis}}
\equalcont{These authors contributed equally}

\author[1,2,3]{\fnm{Daniel} \sur{Rueckert}}
\equalcont{These authors contributed equally}

\author[3]{\fnm{Ben} \sur{Glocker}}
\equalcont{These authors contributed equally}

\affil[1]{\orgdiv{Chair for AI in Healthcare and Medicine}, \orgname{Technical University of Munich (TUM) and TUM University Hospital}, \orgaddress{\country{Germany}}}

\affil[2]{\orgname{Munich Center for Machine Learning}, \orgaddress{\country{Germany}}}

\affil[3]{\orgdiv{Department of Computing}, \orgname{Imperial College London}, \orgaddress{\country{United Kingdom}}}

\affil[4]{\orgname{Hasso Plattner Institute}, \orgaddress{\country{Germany}}}

\abstract{
Medical AI models hold immense potential to improve patient outcomes.
However, these models are also known to unintentionally memorise individual records from their training datasets \cite{zhang2021understanding, feldman2020neural, zhang2023counterfactual, tonekaboni2025an}.
While such memorisation has been linked to targeted privacy attacks \cite{shokri2017membership, carlini2022membership, knolle2026disparate, carlini2023extractingdiffusion, carlini2021extractinglanguage, nasr2025scalableLLMs}, its consequences for clinical deployment, where patients may be assessed by a model that saw their historical data during training, remain poorly understood.
Here we show that predictions on a patient’s unseen future data can change significantly if a model observed that same patient’s anonymised historical data during training, a phenomenon we term “memorisation bias”.
We demonstrate that this bias exists across diverse data modalities and model architectures, and over prolonged time spans: in some cases, we find that memorisation bias can persist on future records acquired decades after the patient's historical records used for model training.
Moreover, in simulated prospective deployment, we find that memorisation bias has asymmetric effects on the diagnostic accuracy of returning data contributors.
When a patient returned with a \textit{de novo} condition that was absent from their historical records in the model's training dataset, diagnostic sensitivity decreased significantly compared to an otherwise identical model not trained on their historical data.
Conversely, when their health state was unchanged, both sensitivity and specificity were significantly inflated.
Together, our findings reveal a previously uncharacterised risk in medical AI that arises when an AI model is deployed on patients that contributed to the model's training data.
This exposes a shortcoming of current model development practice: the de-identification measures designed to protect patients' privacy make it difficult to identify returning data contributors and exclude them from the AI-assisted interpretation of their own future data.
Effectively mitigating memorisation risks may thus require changes to current model training and deployment protocols.
}
\maketitle

%%%%%%%%%%%%%%%%%%%%%%%%%%%%%%%%%%%%%%%%%%%%%%%%%%%%%%%%%%%%%%%%%%%%%%%%%%%%%%%%%%%%%%%%%%%%%%%%%%%%%%%%%%%%%%%%%%%%%%%%%%%%%%%%%%%%%%%%%%%%%%%%%%%%%%%%%%
\section*{Main}
Artificial intelligence (AI) models are quickly moving from retrospective evaluation into routine use in population screening programs and clinical practice.
For good reason.
Given enough high-quality training data, AI models can often match, and in some cases even exceed, the performance of medical experts for diagnostic tasks \cite{esteva2017dermatologist, de2018clinically, mckinney2020international, mcduff2025towards, Brodeur2026LLMreasoningphysician} and prospective trials now demonstrate clinical benefit at scale \cite{gommers2026Masai, eisemann2025nationwide, yao2021artificial}.
Most of these advances are enabled by large-scale training datasets of real patient data.

Training data, however, is not only absorbed in aggregate.
Besides learning general patterns, AI models also retain information about individual records in their training data: a model's output for a given input can depend on whether that same input was present in its training dataset.
This phenomenon, termed memorisation, is well documented across model families, from deep learning models \cite{zhang2021understanding, feldman2020neural, zhang2023counterfactual, carlini2022membership, knolle2026disparate, carlini2023extractingdiffusion, tonekaboni2025an, shokri2017membership} to classical machine learning models \cite{bartlett1998boostingtheMargin, schapire2013explainingAdaboost, wyner2017explainingInterpolating, Zarifzadeh2024lowcost}.
Indeed, research by \citet{feldman2020does} and \citet{feldman2020neural} suggests that such memorisation is not merely an artefact of over-parameterisation, but is in fact required to achieve optimal generalisation performance on the long-tailed data distributions common to many real-world settings.

Previous research has studied memorisation almost exclusively in the context of privacy attacks, where an adversary aims to extract sensitive information from a trained model
\cite{shokri2017membership, carlini2022membership, knolle2026disparate, carlini2023extractingdiffusion, carlini2021extractinglanguage, nasr2025scalableLLMs, tonekaboni2025an}.
As a result, the consequences for patients whose records were memorised and who may encounter the same model again during their future care journey remain largely unexplored.
With this study, we show that when models are deployed prospectively, memorisation creates a specific and under-appreciated risk.
Because a patient's records are highly self-similar over time, a future record may act as a partial cue for a memorised historical record, shifting the model's predictions towards a patient's previous health state.

This is not a hypothetical scenario.
Medical AI models are routinely deployed on the same population from which their training data were sourced.
For example, Germany's national breast cancer screening program uses an AI model trained on $1.2$ million mammograms sourced from the same screening population it now serves \cite{eisemann2025nationwide, leibig2022combining}.
Comparable deployment is underway elsewhere, including national breast cancer screening programmes in the United Kingdom \cite{edith_trial} and Sweden \cite{gommers2026Masai}.
Regulatory guidance actively encourages this, with the EU AI Act \cite{eu2024aiact} requiring that training data reflect the geographical and contextual setting of intended use (Art. 10(4)), and similar international guidelines for medical AI development \cite{imdrf2025gmlp} calling for the intended patient population to be sufficiently represented in a model's training dataset.
If memorisation were present, the risk for harm in AI-assisted population screening programs is substantial: hundreds of thousands of patients will return at regular intervals over years or decades, meaning that data contributors will repeatedly encounter a model that may have memorised their earlier, typically healthy, records.
Whether training data memorisation could systematically alter predictions on data contributors' unseen future data encountered only during prospective deployment has not been systematically studied.
Our study addresses this gap, marking the first time that longitudinal memorisation effects have been studied using real-world patient data.

Here we show that medical AI models, trained to perform standard diagnostic (supervised classification) tasks, exhibit a phenomenon we term \say{memorisation bias}: a systematic shift in a patient's future predictions towards the health state that the model observed in the same patient's historical records during training.
More specifically, using four large datasets spanning electrocardiograms, chest radiographs and electronic health records, we show that including a patient's historical data in a training dataset can significantly alter the model's future predictions for that same patient.
Crucially, while memorisation bias becomes both rarer and weaker over time, we show that it can persist for multiple decades, affecting predictions on records acquired long after the historical data used for model training.
We further show that memorisation bias can lead to systematic diagnostic errors. 
In simulated deployment scenarios, diagnostic sensitivity was significantly reduced for patients who returned with a condition not present in their historical data used to train the model.
On the other hand, sensitivity and specificity were significantly artificially inflated for patients whose health status was unchanged compared to their historical training records.

Our findings pose an immediate and practical dilemma.
Patients are rarely informed that their personal data was used to train a specific model.
This is because model training constitutes a secondary use of health data in most jurisdictions, and typically proceeds under broad consent or a consent waiver rather than study-specific consent.
Moreover, because of standard de‑identification procedures, it is not straightforward to determine whether any given patient is represented in a model's training dataset once the model is deployed.
Protecting data contributors from memorisation bias by excluding them from the AI-assisted interpretation of their future data is thus difficult in practice and also raises ethical questions about disparities in the provision of care.
%As a result, effectively mitigating memorisation risks while equally distributing AI benefits will require changes to current model training and deployment protocols.

\begin{figure}[t]
    \centering
    \includegraphics[width=0.55\linewidth]{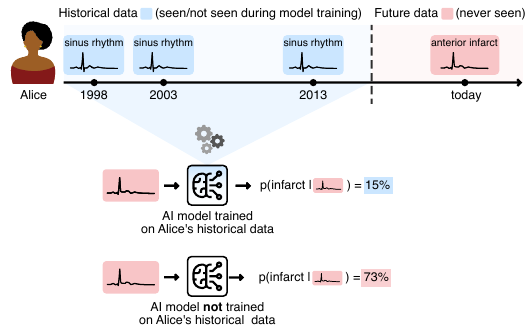}
    \vspace{3mm}
    \caption{\textbf{Schematic of our longitudinal memorisation detection approach.}
    Each patient's records are split temporally: earlier (historical) records may be used for model training; later (future) records are used for model evaluation only and are never used for training or model selection.
    To detect longitudinal memorisation, we compare the predictions models make on a future record when trained on that patient's historical records with those made by models trained without them.
    In the illustrated example, Alice contributed three historical electrocardiograms to the training dataset, all showing a sinus rhythm (a normal heart rhythm), and returns today with an anterior infarct (a heart attack involving the anterior wall of the heart). 
    The model that saw her healthy historical records during training assigns a substantially lower probability to her anterior infarct, potentially leading to a missed diagnosis.
    For clarity, one model is shown per group; our analysis uses $M=100$ models per group for each patient (see Methods for details).
    }

    \label{fig:1}
\end{figure}
%%%%%%%%%%%%%%%%%%%%%%%%%%%%%%%%%%%%%%%%%%%%%%%%%%%%%%%%%%%%%%%%%%%%%%%%%%%%%%%%%%%%%%%%%%%%%%%%%%%%%%%%%%%%%%%%%%%%%%%%%%%%%%%%%%%%%%%%%%%%%%%%%%%%%%%%%%
\subsection*{Detecting longitudinal memorisation}
\begin{figure}[pt]
    \centering
    \includegraphics[width=1\textwidth]{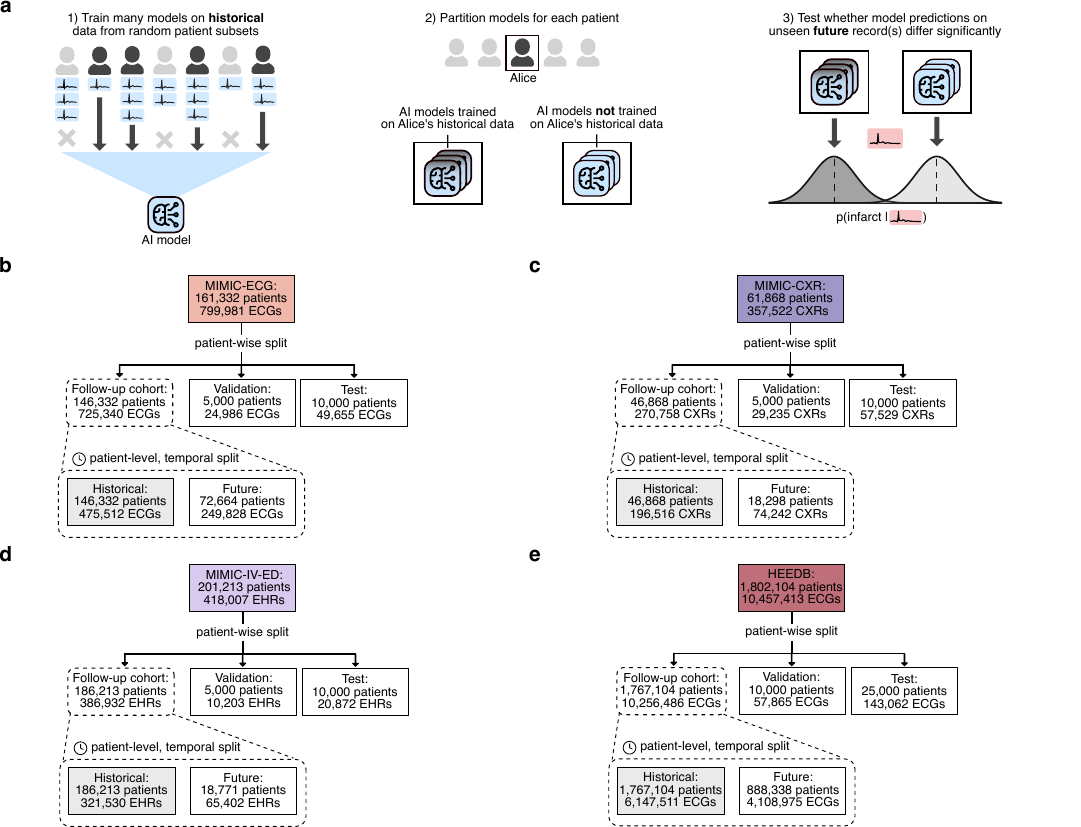}
    \vspace{1mm}
    \caption{\textbf{Study profile}.
    \textbf{a}, the three key steps of our longitudinal memorisation detection approach: (1) training $M=200$ models on the historical data of randomly selected $50\%$ patient subsets, (2) partitioning these models for each patient depending on whether a model's training subset included the respective patient's historical data or not ($M=100$ models per group for each patient), and (3) testing whether the models' predictions on the respective patient's future records differ significantly between model groups.
    \textbf{b-e}, our dataset split strategy.
    For each investigated dataset, we perform a four-way dataset split in two stages:
    (1) a standard, randomised patient-wise split into follow-up cohort, validation dataset and test dataset;
    (2) an additional patient-level, temporal split performed on the records of each patient in the follow-up cohort, yielding the historical and future record datasets.
    Patients with insufficient follow-up data are excluded from the temporal split, and their records are added only to the respective historical dataset.
    By construction, every patient represented in a future dataset also contributes at least one earlier record to the respective historical dataset.
    Sub-panels show resulting split sizes for MIMIC-ECG \cite{Gow2023mimicECG} (b), MIMIC-CXR \cite{johnson2019mimic} (c), MIMIC-IV-ED \cite{xie2022benchmarkingMIMIC-IV-ED} (d) and HEEDB \cite{koscova2026HEEDB}(e).
    Historical data is used for model training (grey boxes with solid lines in \textbf{b-e}), future, validation and test data are used for model evaluation only (white boxes with solid lines in \textbf{b-e}).
    ECGs, CXRs and EHRs refer to electrocardiograms, chest radiographs and electronic health records, respectively.
    }
    \label{fig:2}
\end{figure}

The established gold-standard definition of AI memorisation \cite{feldman2020neural, zhang2023counterfactual} is the marginal change in a model's predictions on an individual record caused by the inclusion (or exclusion) of that same record in its training dataset.
In practice, measuring memorisation involves training a large number of models on random data subsets and comparing the predictions of models whose training subsets included the record of interest with those that did not.
Inspired by prior research on AI memorisation \cite{feldman2020neural, zhang2023counterfactual} and closely related research on membership inference attacks \cite{shokri2017membership, carlini2022membership}, we propose a novel framework to detect memorisation bias.
Specifically, our framework allows us to quantify whether, and to what extent, the inclusion of a patient's historical records in a model training dataset causes a significant change in their future predictions.

Our framework operates as follows (see Fig. \ref{fig:1} \& Fig. \ref{fig:2}a for an overview).
First, we temporally split the data of patients in each follow-up cohort into historical and future records (see Fig. \ref{fig:2}b-e for the resulting split sizes and the Methods for details of the split strategy which enriches the future datasets for \textit{de novo} cases, i.e.\@, positive cases of conditions not present in the respective patient's historical data).
Second, we train $M=200$ models, each on historical data from a randomly selected $50\%$ subset of patients.
Note that these models were trained using state-of-the-art techniques (learning rate schedules, exponential moving parameter average and, where applicable, data augmentation) and using hyperparameters chosen to yield optimal generalisation performance on the validation set; future data was never used for model training or model selection.
Third, for each patient of interest, we partition the models into two groups: those whose training subsets included that patient's historical data and those whose training subsets did not.
Notably, because all models are trained on random data subsets comprising roughly half of the available historical data, performance differences on held-out validation and test sets between models are negligible (Extended Data Fig. \ref{fig:edfig1}a), isolating patient inclusion as the variable of interest.
Using these model groups, partitioned by whether each patient was included in their training data subset, we generate predictions for patients' future records (predicted probabilities for all available annotations/classes).
\citet{feldman2020neural} show that the influence of other patients' data, included or excluded by chance here, is marginalised out quickly and is thus negligible across the large number of models we study.
Fourth, we apply energy-based hypothesis testing to assess whether the two groups of predictions differ significantly (see Methods for details).
A statistically significant difference in the prediction for a patient's future record indicates that the patient's historical data was memorised during model training and that this memorisation influenced the prediction for one of their future records.

%%%%%%%%%%%%%%%%%%%%%%%%%%%%%%%%%%%%%%%%%%%%%%%%%%%%%%%%%%%%%%%%%%%%%%%%%%%%%%%%%%%%%%%%%%%%%%%%%%%%%%%%%%%%%%%%%%%%%%%%%%%%%%%%%%%%%%%%%%%%%%%%%%%%%%%%%%
\subsection*{Memorisation alters future predictions}

\begin{figure}[hp]
    \centering
    \includegraphics[width=1\textwidth]{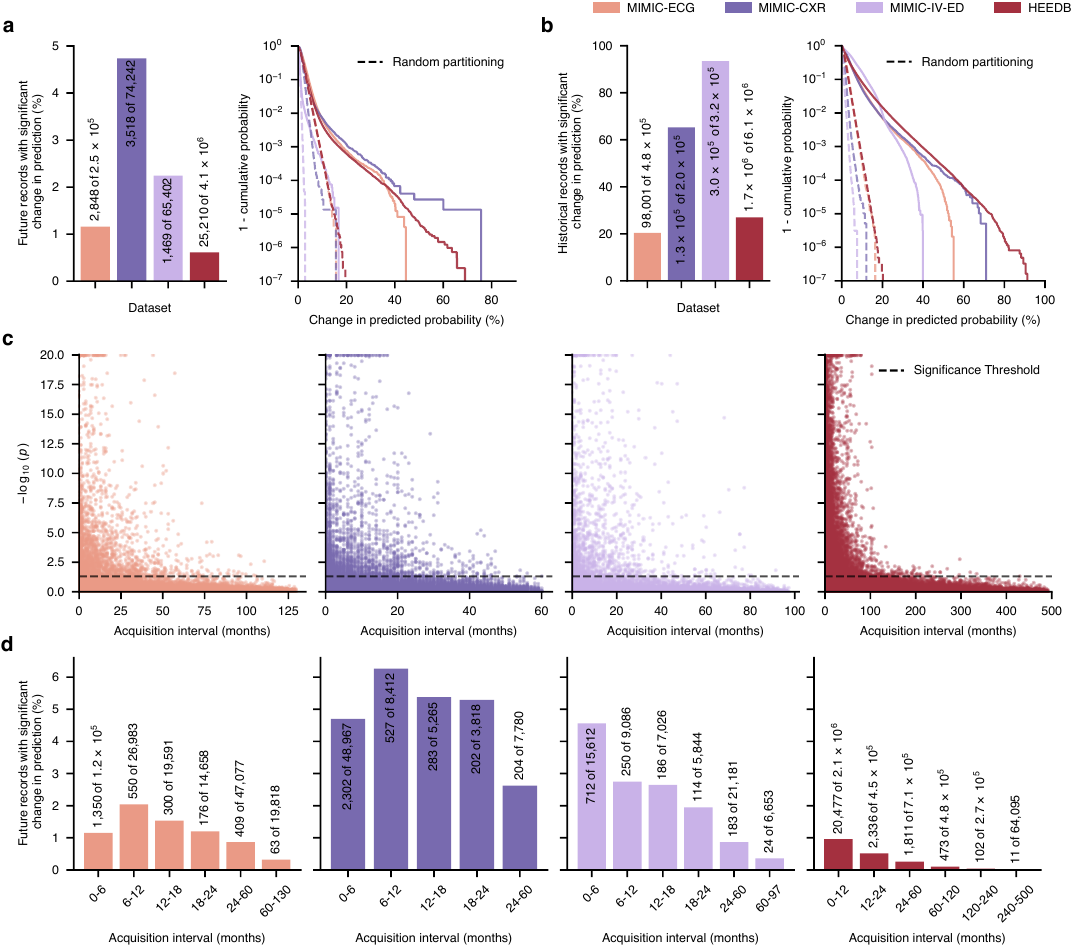}
    \vspace{2mm}
    \caption{\textbf{Historical training data alters future predictions.}
    \textbf{a-b}, the relative share of records for which we detected a significant change in predicted probability due to the patient's inclusion in the training data set (left), alongside empirical complementary cumulative distribution function (ecCDF) plots of effect sizes (right).
    Effect size was measured as the absolute difference in average predicted probability between models trained vs.\@ not-trained on a given patient's historical data (maximum across available classes).
    Panels show results for unseen future records (a) and historical records used for model training (b).
    \textbf{c}, Manhattan plots of multiple-comparison corrected $p$ values versus the time in months between the acquisition of the patient's most recent historical training record and the acquisition of the respective future record (hereafter, the acquisition interval).
    \textbf{d}, relative share of future records with a significant change in predicted probability, binned by the acquisition interval.
    Statistical significance was determined using two-sided, nonparametric, energy-based hypothesis tests comparing the predictions (predicted probabilities across all available classes) of models trained versus not trained on the respective patient's historical data ($M=100$ models per group for each patient).
    The threshold for statistical significance was $p\leq 0.05$.
    Multiple-comparison correction was performed separately across each dataset's historical/future records using the Benjamini-Hochberg method to control the false discovery rate at $\alpha=0.05$; $p$ values were clipped to a minimum value of $10^{-20}$ to improve visibility.
    }
    \label{fig:fig3}
\end{figure}

Across all investigated datasets, we find evidence for memorisation bias.
Specifically, we find that a substantial share of future records show significant changes in their predictions when the respective patient's historical data is included in the training dataset (Fig. \ref{fig:fig3}a, left). 
We also measured the effect size of these prediction changes, i.e.\@, the absolute difference in average predicted probability between models trained vs.\@ not trained on a given patient's historical data.
Notably, while effect sizes are small on average, some records show large changes in their predicted probability for one of the available classes, in some cases over $70$ percentage points.
This is indicated by empirical complementary cumulative distribution function (ecCDF) plots of the effect sizes for all future records (Fig. \ref{fig:fig3}a right).
For a given percentage, the ecCDF plots show the share of records which show a memorisation-induced change in their average predicted probability of at least that percentage or higher.
Corresponding energy test statistics are reported in Extended Data Fig. \ref{fig:edfig1}b.
As a baseline comparison, we provide the same effect size measures and energy test statistics for models that were partitioned randomly (dashed lines in Fig. \ref{fig:fig3}a right and Extended Data Fig. \ref{fig:edfig1}b).
Crucially, across all investigated datasets, these randomly partitioned models yielded zero future records with significant differences in their predictions.

To understand the root cause of these memorisation effects, we performed the same analysis on patients' historical records (the data observed during training and thus directly responsible for any memorisation).
As expected, memorisation effects are substantially stronger here.
Compared to future records, a larger share of historical records show significant changes in their predictions when the respective patient was included in the training dataset (Fig. \ref{fig:fig3}b left), and effect sizes are also correspondingly larger (Fig. \ref{fig:fig3}b right).
This is consistent with the intuition that memorisation is strongest on the data a model directly observed during training, and attenuates -- but, as shown previously, can persist -- when the model encounters a data contributor's unseen future data.
As before, randomly partitioned models yielded zero historical records with significant differences in their predictions across all datasets (corresponding effect sizes and test statistics are reported in Fig. \ref{fig:fig3}b and Extended Data Fig. \ref{fig:edfig1}b, respectively).
This confirms that the observed effects are driven by patient-specific memorisation rather than by random variation in predictions between model groups.
Notably, the memorisation-induced prediction changes on patients' (historical) training records that we quantify here are the signal that a membership inference attack exploits to determine whether a given record was used for model training \cite{shokri2017membership, carlini2022membership, knolle2026disparate}.
Our results demonstrate that this membership signal can extend beyond the historical training records themselves and systematically alters predictions on contributors' unseen future data.

For MIMIC-IV-ED \cite{johnson2021mimicIV, xie2022benchmarkingMIMIC-IV-ED}, a tabular dataset of electronic health records, the results reported above were obtained using a random forest model.
To assess how tabular data memorisation varies for different types of classical machine learning and modern AI models, we also applied our memorisation detection framework to logistic regression and tabular ResNet models \cite{gorishniy2021revisiting}(Extended Data Fig. \ref{fig:edfig1}d-f).
Note that, as before, model and optimisation hyperparameters were chosen to yield optimal generalisation performance on the validation set.
The share of records for which we detected memorisation varied by orders of magnitude between model types: on unseen future records, we detected significant changes in predicted probability for $2.25\%$ ($1{,}469/65{,}402$) of records for the random forest, compared to $0.04\%$ ($25/65{,}402$) for the logistic regression model and $0.12\%$ ($83/65{,}402$) for the tabular ResNet, with corresponding rates of $93.53\%$, $0.77\%$ and $3.61\%$ on historical training records.
Effect sizes reached up to $16.7$ percentage points on future records for the random forest, compared to $1.2$ and $10.2$ percentage points for the logistic regression model and the tabular ResNet, respectively.
Together, these results suggest that memorisation bias is not limited to modern AI models and that random forests, a model class widely used for tabular clinical prediction problems \cite{abdulazeem2023systematic}, are particularly prone to memorisation bias.

\subsection*{Memorisation can persist for decades}
Next, we asked how long memorisation bias can persist for.
To investigate this, we closely examined future records for which we detected memorisation and, for each, measured the time interval between the acquisition of the patient's most recent historical training record and the acquisition of the respective future record.
Plots of multiple-comparison-corrected $p$ values from our per-record test for memorisation bias against this acquisition interval (Fig. \ref{fig:fig3}c) reveal that memorisation bias can persist over remarkably long time spans.
For example, for HEEDB \cite{koscova2026HEEDB}, a large dataset comprising electrocardiograms from $1.8$ million patients collected from the 1980s until 2025, we detect statistically significant changes in model predictions for future records acquired more than 25 years after the patient's most recent historical training record.

To further characterise how memorisation decays over time, we computed the relative share of memorised future records as a function of this acquisition interval (Fig. \ref{fig:fig3}d).
The rate at which future records exhibit memorisation bias, as well as the decay of this rate for increasing acquisition intervals, differs considerably between datasets.
We hypothesise that this reflects differences in how strongly a patient's future records resemble their historical records and natural differences in the intervals at which patients return.
A survival-style analysis of effect sizes for increasing acquisition intervals is reported in Extended Data Fig. \ref{fig:edfig2}, ecCDF plots of the acquisition intervals over all future records, as well as the subset of future records for which we detected memorisation bias, are reported in Extended Data Fig. \ref{fig:edfig1}c.
Across datasets, we find that memorisation bias generally becomes less pronounced and rarer with longer acquisition intervals.
Nevertheless, we find that predictions on some future records with large acquisition intervals still exhibit significant memorisation-induced prediction changes.
Together, these results suggest that memorisation bias is not a transient artefact that resolves completely as a patient's data changes over time, but a persistent phenomenon capable of affecting predictions decades into the future.
\begin{figure}[pt]
    \centering
    \includegraphics[width=1\textwidth]{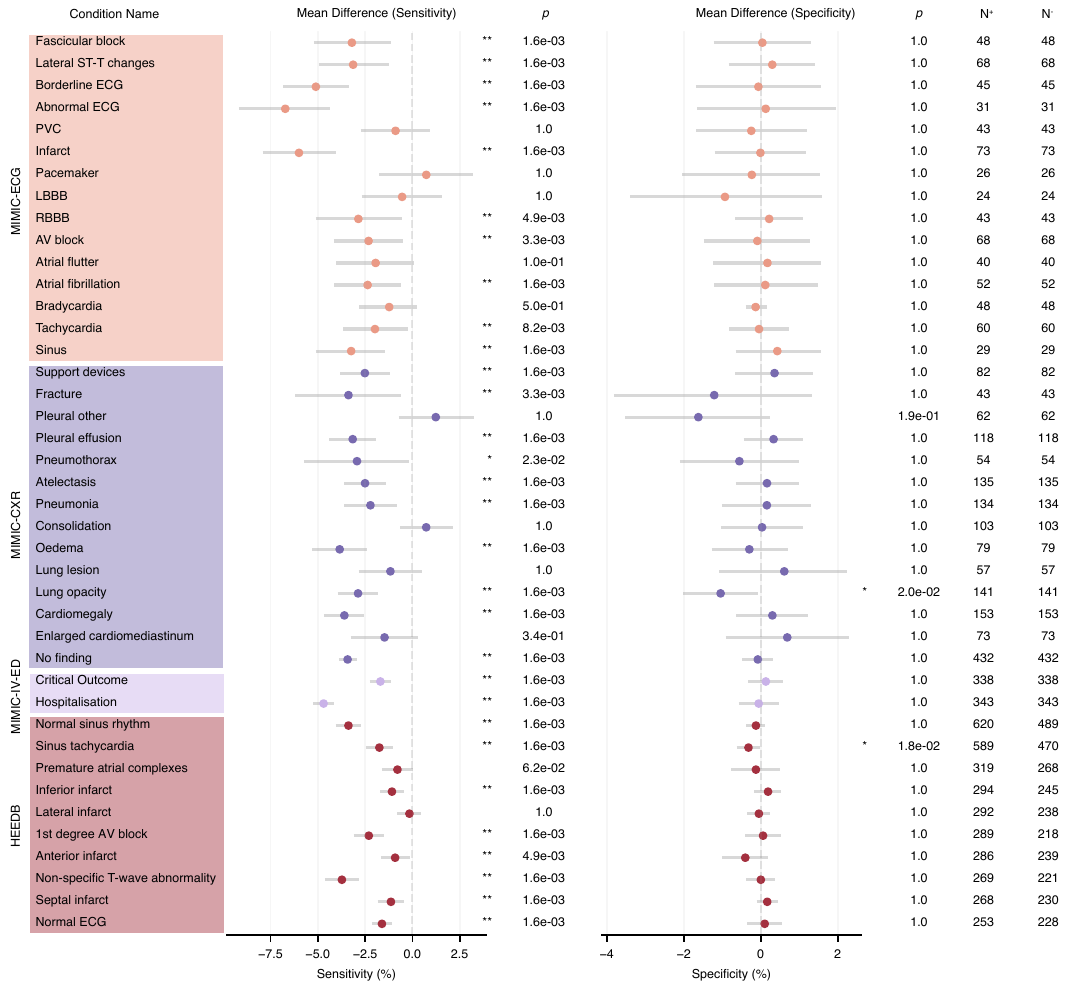}
    \caption{\textbf{Memorisation bias reduces diagnostic sensitivity on future records of patients returning with \textit{de novo} conditions.}
    Memorisation-induced changes in diagnostic sensitivity and specificity on future records of patients returning with a \textit{de novo} condition, i.e.\@ a condition for which no positive cases were present in that patient's historical training record(s).
    Each row represents a single condition: the left-hand panel shows the mean difference in sensitivity between models trained vs.\ not trained on patients' historical data ($M=100$ models per group for each patient), the right-hand panel the corresponding difference in specificity.
    $N^+$ and $N^-$ denote the number of positive and negative cases.
    Evaluated cases comprised patients returning with a \textit{de novo} positive diagnosis, each matched to a randomly selected age- and sex-matched control where possible.
    Condition-specific decision thresholds maximised Youden's $J$ statistic on the test set.
    Statistical significance was determined using exact, two-sided permutation tests that reassign model indices to the corresponding patient subset masks ($B=100{,}000$ permutations per comparison); error bars denote simultaneous $95\%$ confidence intervals obtained by inverting the same test.
    $p$ values and confidence intervals were corrected across all $C=82$ illustrated comparisons using the Bonferroni method.
    * and ** denote $p\leq0.05$ and $p\leq0.01$, respectively; the resolution of the permutation test floors raw $p$ values at $2/(B+1)\approx2\times10^{-5}$, and hence corrected $p$ values at $1.6\times10^{-3}$.
    For HEEDB, we present results only for the $10$ conditions with the largest number of \textit{de novo} cases; data for all conditions are reported in Extended Data Fig.~\ref{fig:edfig4}.
    }
    \label{fig:fig4}
\end{figure}

\subsection*{Memorisation affects diagnostic accuracy}
After establishing that including a patient's historical data in a model's training dataset can systematically alter their future predictions, we next investigated whether these changes could have diagnostic implications.
To this end, we simulated the prospective deployment of AI models on patient populations that contributed to their training data and quantified the impact of memorisation bias on their diagnostic accuracy.

First, we determined optimal decision thresholds for the annotated conditions in each investigated dataset by maximising Youden's $J$ statistic \cite{youden1950index, peirce1884numerical} on the respective held-out test set.
To remove patient-level inclusion effects and general random variation across models, this was performed with predictions averaged across all random-subset models for each dataset.
Second, using these thresholds, we computed the corresponding clinical decisions for patients' future record predictions from models trained vs not trained on each patient's historical data.
Third, we compared the sensitivity and specificity of these decisions.
Notably, we stratified patients' future records into two groups: future records where patients returned with a \textit{de novo} condition -- a positive case of a condition that was not present in any of their historical training records -- and future records where patients returned with unchanged health states (comprising both positive and negative cases).
Since the \textit{de novo} group comprises only positive cases, we sampled an equal number of age- and sex-matched controls for each condition to enable standard sensitivity and specificity analyses.

The results reveal a striking asymmetry: on future records of patients returning with \textit{de novo} conditions, models trained on patients' historical data showed significantly lower sensitivity than models not trained on that data (Fig. \ref{fig:fig4}).
Conversely, on future records of patients returning with a health state unchanged relative to their historical training records, models trained on patients' historical data showed significantly increased sensitivity and specificity compared to models not trained on that data (Extended Data Fig. \ref{fig:edfig3}), artificially inflating apparent diagnostic performance.
In other words, memorisation bias is associated with a higher rate of false-negative model predictions when contributors return with a condition that was absent from their historical training records.
When contributors return with unchanged health states, memorisation bias artificially increases both: sensitivity and specificity.

\subsection*{Risk mitigation via differential privacy}
\begin{figure}[!htp]
    \centering
    \includegraphics[width=0.9\linewidth]{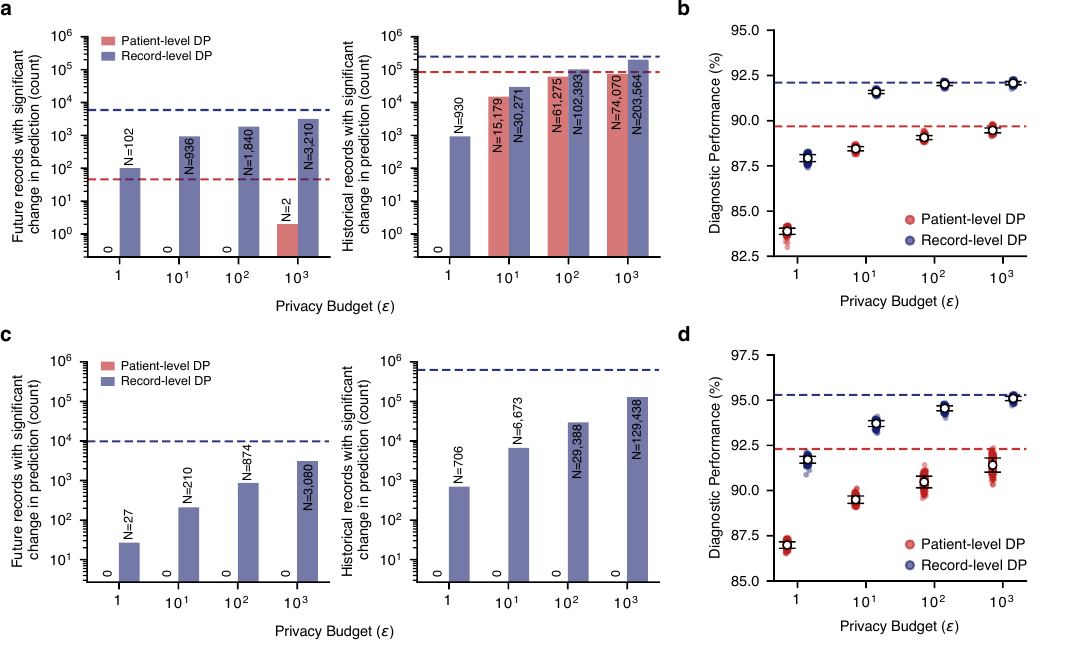}
    \caption{\textbf{Differential privacy reduces memorisation.}
    \textbf{a-b} Results for MIMIC-ECG. 
    \textbf{a}, counts of records with a significant change in prediction due to patient inclusion for models trained with decreasing levels of ($\varepsilon$, $\delta$)-DP privacy protection (increasing $\varepsilon$ values).
    Panels show results for unseen future records (left) and historical training records (right).
    $\delta$ was kept constant at $1/D$ where $D$ is size of the respective historical training data subset.
    \textbf{b}, diagnostic performance of the respective models on the unseen test set ($M=200$ models for each setting); coloured points, macro average area under the receiver operating characteristic curve scores (AUROC, average across all available classes) of individual models; white markers with error bars denote mean $\pm$ s.d.
    \textbf{c-d}, as in panels (\textbf{a-b}) but for HEEDB.
    Statistical significance was determined using two-sided, multivariate, nonparametric energy-based hypothesis tests comparing the predicted probabilities of models trained vs.\@ not trained on the respective patient's historical data ($M=100$ models per group for each patient).
    Multiple comparison correction was performed separately for each experimental setting's historical/future records using the Benjamini-Hochberg method to control the false discovery rate at $\alpha=0.05$.
    Patient-level DP protection was achieved by discarding all but the most recent historical training record per patient and then applying record-level DP accounting; the future record dataset was not modified.
    Dashed lines indicate results for non-private baseline models, which were trained with the same data, hyperparameters and training recipe as the private models.
    }
    \label{fig:fig5}
\end{figure}

Next, we investigated which level of differential privacy \cite{dwork2014algorithmic} (DP) protection would be needed to prevent memorisation bias.
When applied during model training \cite{abadi2016deep}, DP provably limits the amount of information a model can extract from an individual training record and thereby naturally reduces memorisation.
Notably, DP can also be implemented at the patient-level, where it instead provides an upper bound on the information extracted from all records that a single patient contributes to a training dataset.
We trained a range of MIMIC-ECG \cite{Gow2023mimicECG} and HEEDB models with increasing levels of record- and patient-level ($\varepsilon$, $\delta$)-DP privacy protection.
Owing to computational resource constraints, the model size had to be reduced from ViT-S-25 (14 million parameters) to ViT-T-25 (1.9 million parameters); we therefore trained separate non-private baselines for each accounting regime, matched to the corresponding private models in architecture, data, hyperparameters and training recipe.
As expected, we find that memorisation generally decreases with increasing levels of DP privacy protection (i.e.\@ smaller $\varepsilon$ values; Fig. \ref{fig:fig5}a\&c), at the cost of reduced diagnostic performance on the held-out test set (Fig. \ref{fig:fig5}b\&d).

However, this expected privacy-utility trade-off \cite{dwork2014algorithmic} conceals a more consequential distinction between the two accounting regimes.
Record-level DP, the variant typically used across research and industry for its simplicity, substantially reduced but did not eliminate memorisation bias.
Even at the strongest privacy budget tested ($\varepsilon=1$), we still detected significant memorisation-induced prediction changes in a substantial share of historical and future records (Fig. \ref{fig:fig5}a\&c), with effect sizes on future records at the largest budgets reaching over $40$ percentage points for both datasets (Extended Data Fig. \ref{fig:edfig6}).
These findings match results on patient-level membership inference attack susceptibility under record-level DP protection reported by \citet{knolle2026disparate}.

By contrast, models trained with patient-level DP protection showed almost no signs of memorisation bias.
Although we still detected memorisation on patients' historical records at weaker privacy budgets for MIMIC-ECG (up to $N=74,070$ records at $\varepsilon=10^3$), this memorisation essentially did not transfer to patients' future records, where we detected at most two affected records across all tested budgets ($N=0,0,0,2$ for $\varepsilon=1,10^1,10^2,10^3$; Fig. \ref{fig:fig5}a).
For the larger HEEDB dataset, patient-level DP protection yielded no significant changes in predictions across all tested budgets, in either the historical or future records.
Correspondingly, ecCDF plots of energy test statistics and memorisation effect sizes on future records for $\varepsilon\leq10^3$ patient-level DP do not deviate substantially from random partitioning baselines (Extended Data Fig. \ref{fig:edfig6}).
Together, these findings suggest that DP, implemented at the patient-level, is much more effective at preventing memorisation bias than its record-level counterpart.

%%%%%%%%%%%%%%%%%%%%%%%%%%%%%%%%%%%%%%%%%%%%%%%%%%%%%%%%%%%%%%%%%%%%%%%%%%%%%%%%%%%%%%%%%%%%%%%%%%%%%%%%%%%%%%%%%%%%%%%%%%%%%%%%%%%%%%%%%%%%%%%%%%%%%%%%%%
\subsection*{Discussion}
We present data from the first investigation into longitudinal AI memorisation, building on our preliminary, earlier work \cite{knolle2025memorisation}.
Using simulation experiments grounded in real-world longitudinal patient data, we provide early evidence that AI memorisation could translate into downstream diagnostic harm for data contributors.
Together, our results indicate that patients whose historical records were used to train an AI model may face an elevated risk of missed diagnoses when they encounter the same model in their future care journey.

The number of missed diagnoses attributable to memorisation bias in our simulated deployment experiment is modest.
Interpreting this result requires care: \textit{de novo} cases are rare in routinely collected data, which limits their absolute count in the datasets we investigate, while our temporal split strategy deliberately enriched for such cases.
Our estimates should therefore be read as characterising the existence and direction of the effect rather than its expected incidence in deployment.
There is reason to expect the number of missed diagnoses attributable to memorisation bias to increase in the future, although we do not test this directly.
Prior research has shown that the proportion of training data a model memorises increases with model capacity \cite{carlini2021extractinglanguage, carlini2022membership, carliniquantifying, nasr2025scalableLLMs, knolle2026disparate}.
This is concerning, given that current AI model development is guided by \say{scaling laws} \cite{kaplan2020scaling}, which drive rapid growth in both model and dataset sizes in pursuit of improved performance.
As larger AI models are trained on historical data sourced from ever-larger patient populations, the absolute number of individuals affected by memorisation bias is likely to rise drastically, exacerbating the risks we identify here.

Our findings have immediate and far-reaching implications.
Anonymisation has long been assumed to protect data contributors from harm, but our results show that this may no longer hold true in the era of medical AI.
Anonymisation cannot and will not protect contributors against future diagnostic errors caused by memorisation bias.
Worse, it may even obscure who is at risk.
Because AI training datasets are routinely anonymised, neither patients, nor model developers, nor healthcare providers can easily tell who is affected once a model is deployed.
Given current deployment practices and infrastructure, it may thus not be straightforward to protect data contributors by excluding them from the AI-assisted interpretation of their own future data.
This suggests that the common practice of building medical AI models using anonymised, historical patient data needs a fundamental reassessment and motivates the search for principled risk-mitigation techniques.

Effectively mitigating risks from memorisation bias will require concerted efforts by model developers, independent researchers, regulatory authorities and other key stakeholders.
This is because the most intuitive mitigation strategies each raise practical difficulties.
The simplest option would be to establish new communication channels between training dataset curators and healthcare providers, enabling the exclusion of data contributors from the AI-assisted interpretation of their future data.
Such channels, however, could pose privacy risks by revealing membership information \cite{shokri2017membership, knolle2026disparate} and run counter to the purpose of the de-identification procedures currently in use.
Exclusion could also be made more targeted by using membership inference attacks as memorisation detectors, flagging patients whose records appear to have been memorised during training.
This, too, comes with practical issues: membership inference attacks are computationally expensive to evaluate at patient-level resolution \cite{knolle2026disparate}, which would be required, as a missed detection leaves a patient unprotected.
Underlying all of these approaches is the assumption that patients whose data were not used for model training still exist.
Current trends in model development based on scaling laws \cite{kaplan2020scaling} suggest this may not hold in the future.
As a model's training dataset size approaches population coverage, exclusion becomes self-defeating: withholding a model from a substantial share of a patient population limits its practical usefulness and raises ethical questions about disparities in the provision of care.

DP \cite{dwork2014algorithmic} could be one of potentially multiple risk mitigation strategies that do not require excluding data contributors from prospective model deployment, but it requires careful implementation.
Our results underscore that the choice of privacy unit for implementing DP is consequential.
Record-level DP, the variant most commonly used across research and industry for its simplicity, substantially reduced but did not eliminate memorisation bias in our experiments, with significant memorisation effects still detectable even at $\varepsilon=1$, the strongest level of protection we tested.
Patient-level DP, by contrast, was much more effective at preventing memorisation bias.
However, even patient-level DP is not a panacea.
Patients with duplicate entries in a database, or familial and hereditary similarities between distinct patients, can violate the assumption that each protected unit is independent, and thereby weaken effective protection.
Beyond DP, our results suggest that any effective mitigation will need to protect patients rather than records.
Because memorisation bias arises from the self-similarity of a patient's records over time, safeguards applied at the record level leave the underlying vulnerability intact.
Establishing effective patient-level protection at scale, while minimising utility cost, remains an open research problem.

Our study has several limitations.
First, patients with long follow-up are rare in the datasets we investigate, so our estimates of the persistence of memorisation bias at longer acquisition intervals rest on comparatively few records and are thus likely conservative.
Second, our analysis of diagnostic harm is simulated rather than observed.
We estimate diagnostic effects from changes in model predictions under post-hoc thresholds, with models assessed in isolation rather than in a clinician-in-the-loop setting, so the true downstream clinical impact of memorisation bias remains to be quantified prospectively.
Our results also suggest that prospective trials would require careful design: without a \textit{de novo} case stratification, a trial would likely reach incorrect conclusions about diagnostic primary endpoints for data contributors, as patients often return with unchanged health states.
Third, the utility cost we observe for DP is likely an overestimate.
Our patient-level DP results rely on a naive implementation that discards all but one record per patient before applying record-level accounting, so the reduced diagnostic performance on the unseen test data we report reflects this data loss as much as the privacy mechanism itself.
On sufficiently large datasets, patient-level DP approaches that retain multiple records per patient and build on the engineering insights of \citet{de2022unlocking} and \citet{mckenna2025scalingDP} would likely close some of this gap.
Fourth, we did not perform a subgroup analysis.
Prior research \cite{knolle2026disparate} showed that membership inference risk is not distributed equally, with a disproportionate share of the privacy risk burden falling on patient groups underrepresented in the training data; whether these disparities extend to memorisation bias remains an open question.
Fifth, we studied diagnostic models trained for supervised classification tasks.
Prior research on memorisation \cite{carlini2022membership, hayes2025exploringMIAsLLMS, carlini2023extractingdiffusion, wang2024memorizationSSL} suggests that the phenomenon we report in this study should also arise in other settings (such as e.g.\@, models trained for segmentation/survival prediction, or foundation and large language models), but how it manifests in each of them requires further dedicated research.

Together, our findings indicate that including a patient's personal medical data in an AI model's training dataset can systematically alter the model's predictions on the patient's unseen future data, and that these changes are directionally unfavourable when their health state changes.
While the magnitude of the resulting clinical harm remains to be quantified in prospective studies, our findings suggest that memorisation warrants consideration not only as a privacy risk but as a potential source of diagnostic harm concentrated on the very individuals who make medical AI possible.

\subsection*{On the connection to \citet{knolle2026disparate}}
At a purely technical level, we build closely on the findings of \citet{knolle2026disparate} where the authors demonstrated that the predicted probability an AI model assigns to a given record shifts substantially when that same record is included in the model's training dataset.
For some patients, this shift in predicted probability enables near-perfect success rates for membership inference attacks.

With this study, we provide evidence for a closely related phenomenon.
Specifically, we show that predicted probability shifts caused by including a patient's historical records in a model’s training dataset are not limited to those records; they also extend to unseen future records from the same data contributors.
Notably, this includes records in which the patient’s health state has changed, suggesting that, in some cases, models rely on patient-specific rather than disease-specific features to make their predictions.
This finding carries fundamentally different implications from the ones presented by \citet{knolle2026disparate}.
While \citet{knolle2026disparate} focuses on the vulnerability to targeted privacy attacks, our results suggest that using anonymised patient data for AI model training could cause tangible, real-world harm to data contributors in the form of future misdiagnoses.
Crucially, this harm arises naturally from the deployment of models trained on data sourced from ever-larger patient populations and does not require interference from adversarial actors.

\clearpage

\bibliography{bibliography}% common bib file

\clearpage
%TC:ignore
%%%%%%%%%%%%%%%%%%%%%%%%%%%%%%%%%%%%%%%%%%%%%%%%%%%%%%%%%%%%%%%%%%%%%%%%%%%%%%%%%%%%%%%%%%%%%%%%%%%%%%%%%%%%%%%%%%%%%%%%%%%%%%%%%%%%%%%%%%%%%%%%%%%%%%%%%%
\section*{Methods}

\bmhead{MIMIC-ECG processing and model training details}
MIMIC-ECG \cite{Gow2023mimicECG, johnson2021mimicIV} is a large electrocardiography dataset consisting of $799,981$ twelve-lead electrocardiograms (ECGs) collected from $161,332$ patients at the Beth Israel Deaconess Medical Center.
Each ECG is $10$ seconds long and paired with structured diagnostic labels that we derived by regular-expression matching on machine-generated text reports.
Signals were resampled to $250$ Hz and pre-processed using a $50$ Hz notch filter, a bandpass filter between $0.67$ Hz and $40$ Hz, and a median filter to reduce baseline wander.
Signal data were normalised to zero mean and unit variance.
Models for this dataset were trained without data augmentation to perform multi-label classification across the following $15$ classes: sinus rhythm, tachycardia, bradycardia, atrial fibrillation, atrial flutter, atrioventricular block, right bundle branch block (RBBB), left bundle branch block (LBBB), pacemaker, infarct, premature ventricular contraction, abnormal ECG, borderline ECG, lateral ST-T changes and fascicular block.
For model training, we used a modified vision transformer (ViT-S-25) \citeMethods{dosovitskiy2021ViT} where we replaced two-dimensional convolution layers with their one-dimensional counterparts.
On the unseen test dataset ($N=49,655$ records), a ViT-S-25 with around 14 million parameters, trained on historical records from a randomly selected $50\%$ of patients, achieves a macro-average AUROC of $93.45 \pm 0.18\%$ across all classes.

\bmhead{MIMIC-CXR processing and model training details}
MIMIC-CXR \cite{johnson2019mimic} is a large chest radiograph dataset consisting of $357,522$ chest radiographs from $61,868$ patients at the Beth Israel Deaconess Medical Center.
We utilise MIMIC-CXR-JPG \citeMethods{johnson2019mimicjpg}, a subsequent re-release of the original dataset which contains images in \say{.jpg} format and structured labels derived from free-text radiology reports.
The structured labels indicate the presence or absence of $14$ common thoracic conditions.
For this dataset, we trained Densenet-121 \citeMethods{huang2017densenet} models (pre-trained on ImageNet \citeMethods{deng2009imagenet}, a large natural image dataset) with data augmentation (random horizontal flipping, random pixel shifts, and random rotations) to detect the presence of the following 
$14$ classes: no finding, enlarged cardiomediastinum, cardiomegaly, lung opacity, lung lesion, oedema, consolidation, pneumonia, atelectasis, pneumothorax, pleural effusion, pleural other, fracture, support devices.
On the unseen test dataset ($N=57,529$ records), a DenseNet-121 with around $7$ million parameters trained on the historical records from a random $50\%$ patient subset achieves a macro-average AUROC score of $79.68 \pm 0.21\%$ across all classes.

\bmhead{MIMIC-IV-ED processing and model training details}
MIMIC-IV-ED \cite{johnson2021mimicIV, xie2022benchmarkingMIMIC-IV-ED} is a large electronic health record dataset comprising $418,007$ records from $201,213$ patients at the emergency department of the Beth Israel Deaconess Medical Center.
We followed the pre-processing steps from \cite{xie2022benchmarkingMIMIC-IV-ED} and trained models without data augmentation to predict hospitalisation and critical outcomes using $64$ clinical features.
For this dataset, we trained random forest models using the scikit-learn \citeMethods{scikit-learn} implementation with n\textunderscore estimators=100, min\textunderscore samples\textunderscore leaf=4, and otherwise default parameters.
These model hyperparameters were chosen because they yielded the highest diagnostic performance on the validation dataset.
On the unseen test dataset ($N=20,872$ records), a random forest trained on the historical records from a random $50\%$ patient subset achieves a macro-average AUROC score of $83.39 \pm 0.08\%$ across all classes.

\bmhead{HEEDB processing and model training details}
The Harvard-Emory ECG Database (HEEDB) \cite{koscova2026HEEDB} is a large electrocardiography dataset from the Massachusetts General Hospital and Emory University Hospital.
Because we are interested in patients with long follow-up data, we use only data from the Massachusetts General Hospital, comprising of $10,457,413$ twelve-lead ECGs from $1,802,104$ patients.
Each ECG is 10 seconds long and paired with structured diagnostic labels derived from a semi-automated diagnosis system.
Signals were resampled to a resolution of $250$ Hz and pre-processed using a $50$ Hz notch filter, a bandpass filter between $0.67$ Hz and $40$ Hz and a median filter to reduce baseline wander.
Signal data were normalised to zero mean and unit variance.
Models for this dataset were trained without data augmentation to perform multi-label classification across the following 36 classes: normal ECG, abnormal ECG, normal sinus rhythm, sinus bradycardia, atrial fibrillation, sinus tachycardia, left axis deviation, premature ventricular complexes, borderline ECG, right bundle branch block, septal infarct, non-specific T-wave abnormality, premature atrial complexes, anterior infarct, left bundle branch block, lateral infarct, non-specific ST abnormality, left ventricular hypertrophy, atrial flutter, left anterior fascicular block, right axis deviation, anteroseptal infarct, anterolateral infarct, right atrial enlargement, inferior infarct, bi-fascicular block, left posterior fascicular block, bi-atrial enlargement, 1st degree atrioventricular block, inferior-posterior infarct, supraventricular tachycardia, wide QRS tachycardia, Wolff-Parkinson-White syndrome, acute pericarditis, posterior infarct, bi-ventricular hypertrophy.
For model training, we used a modified vision transformer (ViT-S-25) \citeMethods{dosovitskiy2021ViT} where two-dimensional convolution layers were replaced with their one-dimensional counterparts.
On the unseen test dataset ($N=143,062$ records), a ViT-S-25 with around $14$ million parameters, trained on historical records from a random $50\%$ subset of patients, achieves a macro-average AUROC score of $95.38 \pm 0.22\%$ across all classes.

\bmhead{General model training details}
All models were trained to perform supervised multi-label classification using the AdamW \citeMethods{loshchilov2019AdamW} optimiser, exponential moving parameter average (EMA), weight decay, and a learning rate schedule in form of a cosine decay with linear warm-up.
Optimisation hyperparameter values were determined via a random search to maximise diagnostic performance (macro-average AUROC across all available classes) on unseen validation data.
Specifically, for each dataset, a random search with $T=50$ trials over suitable values for the learning rate, weight decay, momentum and EMA decay was conducted (results are reported in Supplementary Material Table 1).
To prevent overfitting, which is known to exacerbate memorisation \citeMethods{yeom2018overfitting}, model checkpointing based on the validation loss was employed, ensuring that only the model weights with the best generalisation performance were retained after training terminated.
Unless stated otherwise, model performance figures are reported as the mean $\pm$ standard deviation across the $M=200$ models trained for the respective configuration.

\bmhead{Patient-level temporal split strategy}
To obtain the historical and future record datasets, we split the records of each patient in the follow-up cohort at a patient-specific time point.
Since patients often present with already documented conditions in routinely collected data, splitting at a fixed calendar date or at a fixed fraction of each patient's timeline yields future record datasets containing too few \textit{de novo} cases to reliably study the negative impact of memorisation.
We therefore determined the split point adaptively for each patient to enrich the future record datasets for \textit{de novo} cases.
Concretely, we sorted each patient's records chronologically and, at every record, counted how many of the conditions ever documented for that patient had not yet appeared.
As this count is monotonically decreasing over time, we placed the split point just before the last record at which at least $C$ conditions remained undocumented.
By construction, a patient's future records then contain the first documented occurrence of at least $C$ conditions while the historical dataset retains as many of their records as this constraint permits.
Patients whose health state does not change across their records, either because they contribute a single record only or because the same conditions are documented throughout, are not split, and all of their records are assigned to the historical dataset.
For patients with too few distinct health states to yield $C$ \textit{de novo} cases (counting the absence of any documented condition as a state), we reduced $C$ for that patient to the largest value that still leaves at least one record in the historical dataset.
The split, therefore, guarantees that every patient represented in the future record dataset also contributes at least one record to the historical dataset used for model training.
We chose $C$ separately for each investigated dataset to balance the number of historical records available for model training against the number of \textit{de novo} cases available for our simulated deployment experiment.
For MIMIC-ECG, MIMIC-CXR, MIMIC-IV-ED and HEEDB we used $C=1,1,1,2$, respectively.

\bmhead{Statistical testing of prediction changes}
For each record, we tested whether the predictions (vectors of predicted probabilities or all available classes) of models trained on the corresponding patient's historical data differed significantly from those of models not trained on that data, an
approach conceptually related to membership inference.
More specifically, we trained $M = 200$ models on the historical data of randomly selected patient subsets and then partitioned the models by the inclusion/exclusion of each patient in the respective model's training subset.
We then compared the predictions of these two model groups using the energy distance \citeMethods{szekely2004testing}, a nonparametric multivariate two-sample statistic that quantifies distributional divergence based on pairwise Euclidean distances between
and within samples.
For the predictions $\mathbf{u}_1, \ldots, \mathbf{u}_n \in \mathbb{R}^p$ of the $n$ models trained on the patient's data and $\mathbf{v}_1, \ldots, \mathbf{v}_m \in \mathbb{R}^p$ of the $m$ models not trained on it, the energy distance is:
\begin{equation}
    \mathcal{E}_{n,m}(\mathbf{u}, \mathbf{v}) =
    \frac{2}{nm} \sum_{i=1}^{n} \sum_{j=1}^{m} \lVert \mathbf{u}_i - \mathbf{v}_j \rVert
    - \frac{1}{n^2} \sum_{i,j=1}^{n} \lVert \mathbf{u}_i - \mathbf{u}_j \rVert
    - \frac{1}{m^2} \sum_{i,j=1}^{m} \lVert \mathbf{v}_i - \mathbf{v}_j \rVert ,
\end{equation}
where $\lVert \cdot \rVert$ denotes the Euclidean norm, $n + m = M=200$ and $p$ is the number of classes in the respective dataset.
The corresponding population quantity is non-negative and zero if and only if the two underlying distributions are identical.
Note that subsets were drawn using a balanced random subset design following \citet{carlini2022membership}: each patient was assigned to a uniformly random half of the $M$ models, independently of all other patients.
This guarantees $n = m = M/2$ for every patient (and thus all of their respective records), which independent Bernoulli sampling would achieve only in expectation.
Computation was performed using the Energy test implemented in the hyppo package (v0.5.2) \citeMethods{panda2019hyppo}.
Energy test $p$ values were calculated using the fast chi-squared approximation \citeMethods{shen2021chisquare}, which is appropriate at the group sizes used here ($n = m = 100$), and adjusted for multiple comparisons using the Benjamini--Hochberg procedure to control the false discovery rate at $\alpha=0.05$.

\bmhead{Statistical testing of receiver operating point differences}
To translate memorisation-induced prediction changes into clinically meaningful quantities, we compared the sensitivity and specificity of the two model groups (those whose training subset included the respective patient's historical data and those that did not) at a fixed operating point.
Class-specific decision thresholds were determined by maximising Youden's $J$ statistic \cite{youden1950index, peirce1884numerical} on the mean predicted probabilities across all $M$ models for the unseen test dataset, and are therefore independent of the subset assignments subsequently
tested. 
Comparisons were restricted to future records for which we detected memorisation and performed separately for \textit{de novo} conditions, where each positive case was matched to a negative control drawn randomly from the same dataset on sex and $5$-year age band, and for records of patients with unchanged health states.
Every model's predictions were binarised at these thresholds individually, retaining the information carried by all $M$ models rather than collapsing each group into a single averaged prediction.
For each record $j$ we computed the difference $\delta_j$ in the proportion of correct classifications between the two groups (models trained on that patient's historical records minus those not trained on them), and took the mean ($T = \frac{1}{N} \sum_{j=1}^{N} \delta_j$)
across the $N$ records as the test statistic.
Evaluated over positive cases, $T$ is the difference in sensitivity, and over negative cases, the difference in specificity.
Because every patient is assigned to a uniformly random half of the $M$ models independently of all other patients, the assignment matrix is exchangeable in the model index and, under the null hypothesis that inclusion does not alter predictions, independent of the observed classifications.
We therefore obtained an exact randomisation test by permuting model labels against subset assignments over $B=100,000$ permutations.
Permuting entire assignment vectors preserves the correlation between records of the same patient induced by patient-level subsampling, so no further adjustment for clustering was required.
With $b_+$ and $b_-$, the numbers of permuted statistics at least as large and at least as small as the observed $T$, the one-sided $p$ values are:
\begin{equation}
    p_{\pm} = \frac{b_{\pm} + 1}{B + 1}.
\end{equation}
Note the addition of $1$ to the numerator and denominator, which prevents a $p$ value of zero being reported from a finite number of draws \citeMethods{phipson2010permutation}.
All reported $p$ values are two-sided, obtained by doubling the smaller tail as $p = \min\{1, \; 2\min(p_+, p_-)\}$.
Confidence intervals were obtained by inverting the same permutation test at the corrected level applied to the $p$ values: the interval is the set of effects $d$ for which the shifted statistic $T - d$ is not rejected, which at per-comparison level $a$, with $c = \lceil a(B+1)/2 \rceil - 1$, gives $[\,T - T^{*}_{(B-c+1)}, \; T - T^{*}_{(c)}\,]$, where $T^{*}_{(i)}$ denotes the $i$-th smallest permuted statistic.

\bmhead{Tabular data model comparison on MIMIC-IV-ED}
The tabular setting permits model classes that the imaging and signal datasets do not, so we conducted the analysis described previously on MIMIC-IV-ED with three architectures spanning very different inductive biases: the random forest described above, $L_2$-regularised logistic regression
($C = 3 \times 10^{-5}$, lbfgs solver, maximum $5000$ iterations), and a fully connected residual network for tabular inputs \cite{gorishniy2021revisiting} (\say{tabular ResNet}, $5$ residual blocks of width $500$, $3.8$ million parameters, trained for $50$ epochs with AdamW \citeMethods{loshchilov2019AdamW} using a learning rate of $10^{-2}$, weight decay $1.0$ and EMA decay $0.99$).
$M=200$ models were trained per architecture under the identical subset protocol previously described.
On the unseen test dataset ($N=20,872$ records), the random forest, the logistic regression model, and the tabular ResNet achieve macro-average AUROC scores of $83.39 \pm 0.08\%$, $82.50 \pm 0.07\%$ and $83.37 \pm 0.17\%$, respectively.
Under random model partitioning, no historical or future records were flagged for memorisation for any of the three architectures.

\bmhead{Differential privacy model training details}
Models for the differential privacy risk mitigation experiments were trained with DP-Adam \citeMethods{kingma2014adam} \cite{abadi2016deep} using the jax-privacy library \citeMethods{jax-privacy2022github}.
Due to the limited computational resources available to us, model size had to be reduced from ViT-S-25 ($14$ million parameters) to ViT-T-25 ($1.9$ million parameters).
Hyperparameters for these models were determined separately for MIMIMIC-ECG and HEEDB using a fixed privacy budget of $\varepsilon=100$ and using a random search with $T=20$ trials over suitable values for the learning rate and EMA decay parameter (see Supplementary Material Table 1 for resulting values).
Weight decay, dropout or other forms of regularisation were not used.
Under record-level protection, the unit of privacy is the individual record.
For patient-level protection, we retained only the most recent record per patient, so that one training record corresponds to one patient and the guarantee extends from records to individuals; this reduces the training set from $475,512$ to $146,332$ historical records for MIMIC-ECG and from $6,147,511$ to $1,767,104$ for HEEDB.
Models were trained at four privacy budgets, $\varepsilon \in \{1, 10, 100, 1000\}$, with $\delta$ set to the reciprocal of the number of (historical) training records of a given random patient subset and therefore differing between record- and patient-level DP dataset variants: $\delta \simeq 2.08 \times 10^{-6}$ and $1.63 \times 10^{-7}$ for record-level MIMIC-ECG and HEEDB, and $\delta \simeq 6.83 \times 10^{-6}$ and $5.66 \times 10^{-7}$ for the corresponding patient-level variants.
Privacy accounting used the privacy loss distribution accountant implemented in jax-privacy (v2.1.0), with the noise multiplier calibrated to the target $\varepsilon$ before training rather than computed post hoc.
Note that the diagnostic performance of the models could likely be improved by performing a separate hyperparameter search for each privacy budget and integrating the engineering insights of \citet{de2022unlocking} and \citet{mckenna2025scalingDP}.

\bmhead{Computational resources}
Reproducing all of our experiments requires the training of $5,200$ models and approximately $10,000$ GPU-hours on NVIDIA A100 GPUs.
The non-private experiments ($200$ models per dataset; $200$ each for the three MIMIC-IV-ED architectures) account for $1,681$ GPU-hours, of which HEEDB alone accounts for $1,069$ GPU-hours at $5.3$ GPU-hours per model.
The differential privacy risk mitigation experiments ($200$ models for each of the four privacy budgets, at both record- and patient-level protection, for HEEDB and MIMIC-ECG) account for the remaining approximately $8,200$ GPU-hours.
Training the logistic regression and random forest baselines for MIMIC-IV-ED requires approximately 4 CPU-hours.
These figures exclude hyperparameter search (a further ${\sim}250$ GPU-hours) and data preprocessing.
Storing the model outputs required for our analysis requires approximately $2$ TB of disk space.

\bibliographystyleMethods{sn-mathphys-num}
\bibliographyMethods{bibliography}

\bmhead{Data availability}
All datasets used in this study are publicly available for research purposes. See GitHub repository (\url{https://github.com/moritzknolle/memorisation_bias}) for access instructions.

\bmhead{Code availability}
The code is available on GitHub (\url{https://github.com/moritzknolle/memorisation_bias}).

\bmhead{Acknowledgments}
We thank Georg Schmidt and the members of the AIM Lab for their feedback and support.
The authors gratefully acknowledge computational resources provided by the Leibniz Supercomputing Centre of the Bavarian Academy of Sciences and Humanities.

%\bmhead{Funding}
%This work was partially funded by ERC Grant Deep4MI (Grant No. 884622) and the German Research Foundation (Project No. 532139938).
%B.G. received support from the Royal Academy of Engineering as part of his Research Chair in Safe Deployment of Medical Imaging AI.

\backmatter

%\bmhead{Acknowledgements}

%Acknowledgements are not compulsory. Where included they should be brief. Grant or contribution numbers may be acknowledged.

%Please refer to Journal-level guidance for any specific requirements.

\clearpage

\section*{Extended Data}\label{secA1}
\clearpage
\setcounter{figure}{0}

\begin{figure}
    \centering
    \includegraphics[width=\linewidth]{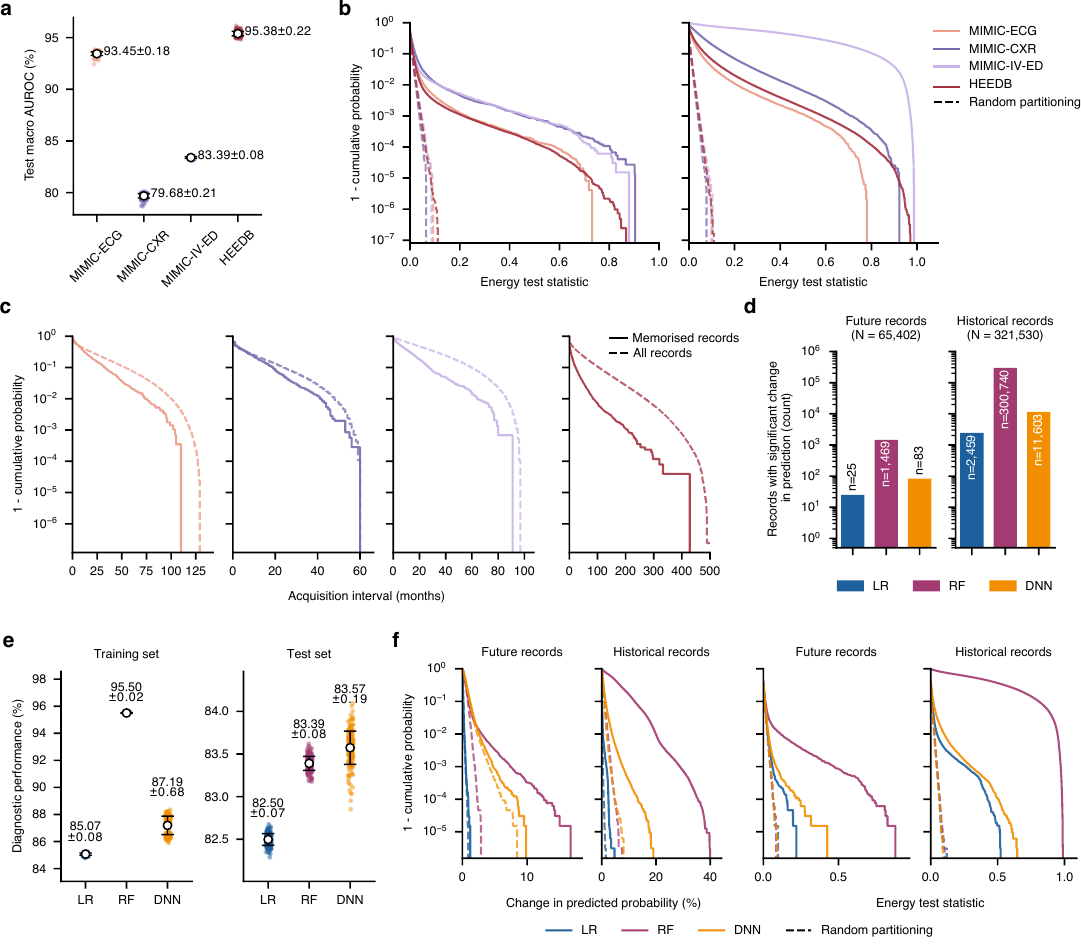}
    \caption{
    \textbf{Additional data on generalisation performance, energy test statistics, acquisition intervals, and memorisation effects for different tabular data architectures.}
    \textbf{a}--\textbf{c}, Results for four datasets: MIMIC-ECG, MIMIC-CXR, MIMIC-IV-ED and HEEDB.
    \textbf{a}, Diagnostic performance of the random subset models on the unseen test sets ($M=200$ models per dataset); coloured points, macro average AUROC (across all available classes) of individual models; white markers with error bars denote mean $\pm$ s.d.
    \textbf{b}, ecCDF of the energy test statistics on unseen future records (left) and historical records used for model training (right).
    All test statistics were derived from energy-based hypothesis tests comparing the two groups' predictions ($M=100$ models per group for each patient).
    \textbf{c}, ecCDF of the acquisition intervals of all future records (dashed lines) and the subset of future records for which we detected memorisation (solid lines).
    \textbf{d}--\textbf{f}, Memorisation across architectures on MIMIC-IV-ED (tabular data): $L_2$-regularised logistic regression (LR), random forest (RF) and tabular ResNet (DNN), each trained with the identical random subset protocol ($M=200$ models each).
    \textbf{d}, Number of records with significantly different predictions (Benjamini--Hochberg FDR correction, $\alpha=0.05$) among future ($N=65{,}402$) and historical ($N=321{,}530$) records.
    \textbf{e}, Diagnostic performance on the training (left) and unseen test set (right), plotted as in \textbf{a}; note the independent y-axes.
    \textbf{f}, ecCDF of effect sizes (left pair) and energy test statistics (right pair), for future and historical records.
    Solid lines in b,f indicate results for models partitioned by patient inclusion/exclusion.
    Dashed lines in b,f indicate random partitioning baselines.
    }

    \label{fig:edfig1}
\end{figure}

\begin{figure}
    \centering
    \includegraphics[width=1.05\linewidth]{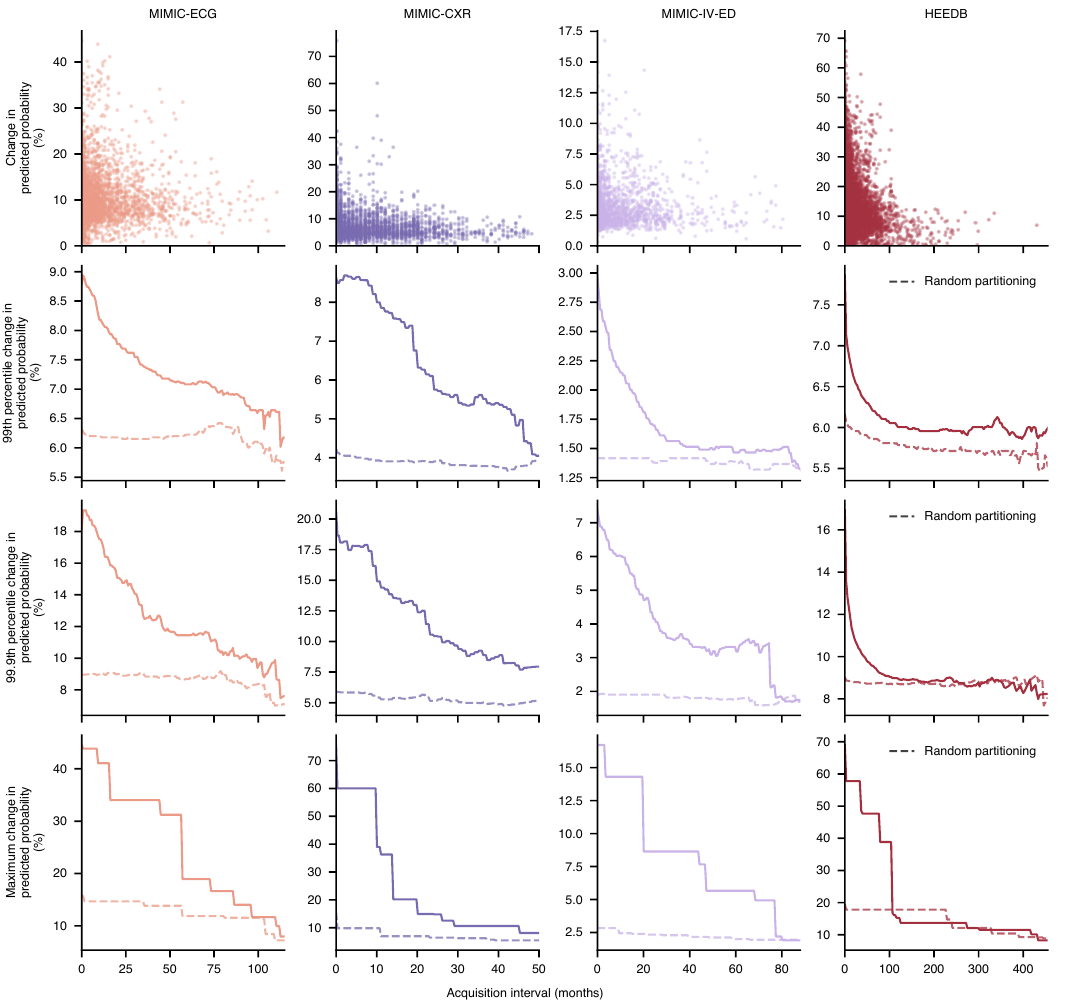}
    \caption{\textbf{Effect sizes and survival-style summary statistics for increasing acquisition intervals.}
    Columns correspond to the four datasets (MIMIC-ECG, MIMIC-CXR, MIMIC-IV-ED, HEEDB).
    The top row is a scatter plot of the effect size against the acquisition interval (in months) for each future record for which a statistically significant change in predicted probability was detected; the remaining rows show survival-style summary statistics of the effect size computed over all future records: 99th percentile (second row), 99.9th percentile (third row), and maximum (bottom row).
    For each future record, the effect size is the maximum absolute difference across diagnostic classes between the mean predicted probabilities of models trained with and without the corresponding patient's historical records (in percentage points).
    In the lower three rows, at each value $X$ on the x-axis, the summary statistic is computed over all records with an acquisition interval of at least $X$ months.
    Points and solid lines indicate the partitioning by patient inclusion/exclusion; dashed lines show identical summary statistics for effect sizes computed from models partitioned at random.
    Within each column, all rows span the same acquisition-interval range, truncated at the largest value beyond which fewer than $N=500$ records remain.}
    \label{fig:edfig2}
\end{figure}

\begin{figure}
    \centering
    \includegraphics[width=\linewidth]{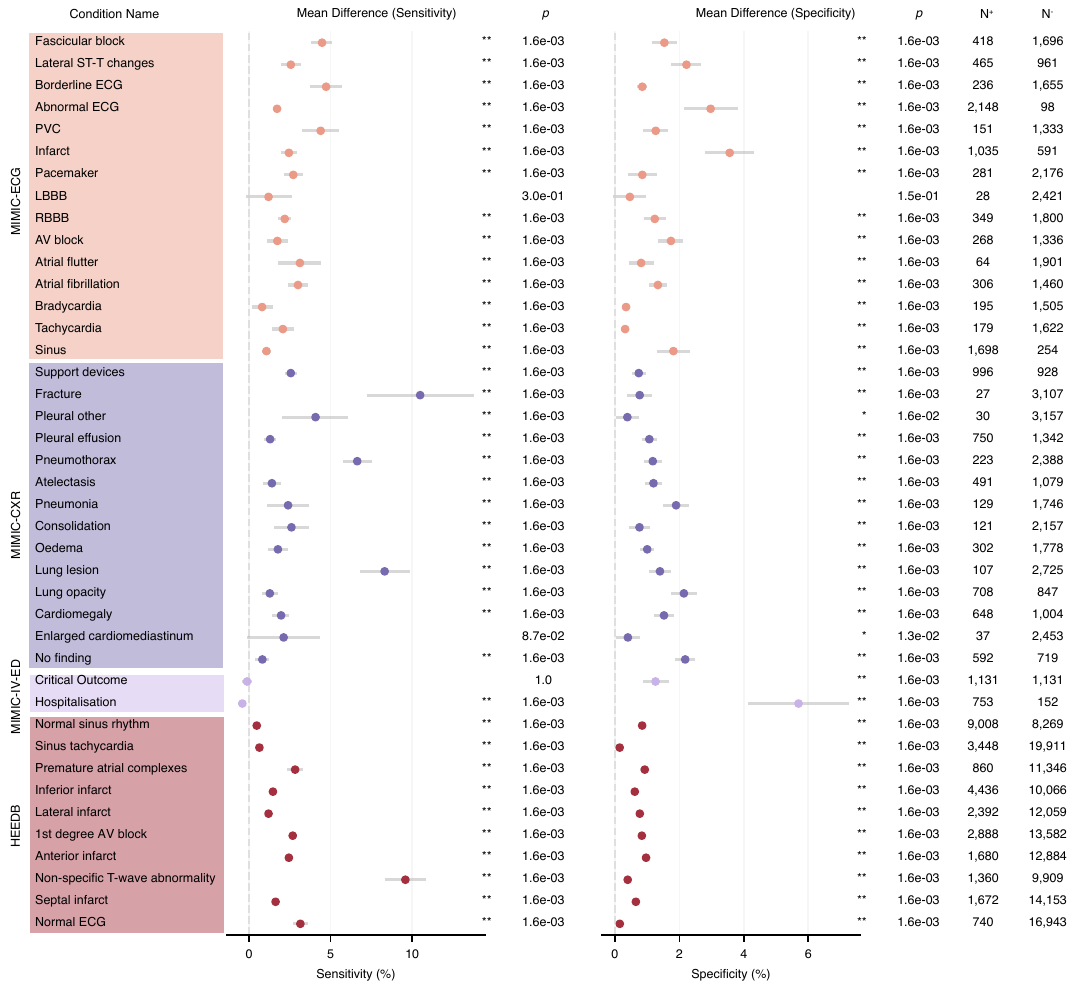}
    \caption{\textbf{Memorisation bias artificially inflates diagnostic sensitivity and specificity on future records of patients returning with unchanged health states.}
    Memorisation-induced changes in diagnostic sensitivity and specificity on future records of patients whose health state was unchanged relative to their historical training records.
    Evaluated cases comprised positive cases of a condition for which at least one positive case was already present in that patient's historical training records, and negative cases of a condition for which only negative historical cases were present.
    Each row represents a single condition: the left-hand panel shows the mean difference in sensitivity between models trained vs.\ not trained on patients' historical data ($M=100$ per group), the right-hand panel the corresponding difference in specificity.
    Point estimates to the right of zero indicate increased performance on data-contributing patients' future records, those to the left a reduction.
    $N^+$ and $N^-$ denote the number of positive and negative cases for each condition, respectively.
    Condition-specific decision thresholds maximised Youden's $J$ statistic on the test set.
    Statistical significance was determined using exact, two-sided permutation tests that reassign model indices to the corresponding patient subset masks ($B=100{,}000$ permutations per comparison); error bars denote simultaneous $95\%$ confidence intervals obtained by inverting the same test.
    $p$ values and confidence intervals were corrected across all $C=82$ shown comparisons using the Bonferroni method.
    * and ** denote $p\leq0.05$ and $p\leq0.01$, respectively; the resolution of the permutation test floors raw $p$ values at $2/(B+1)\approx2\times10^{-5}$, and hence corrected $p$ values at $1.6\times10^{-3}$.
    For HEEDB, we present results only for the $10$ conditions with the largest number of \textit{de novo} cases; data for all conditions are reported in Extended Data Fig.~\ref{fig:edfig5}.
    }
    \label{fig:edfig3}
\end{figure}

\begin{figure}
    \centering
    \includegraphics[width=\linewidth]{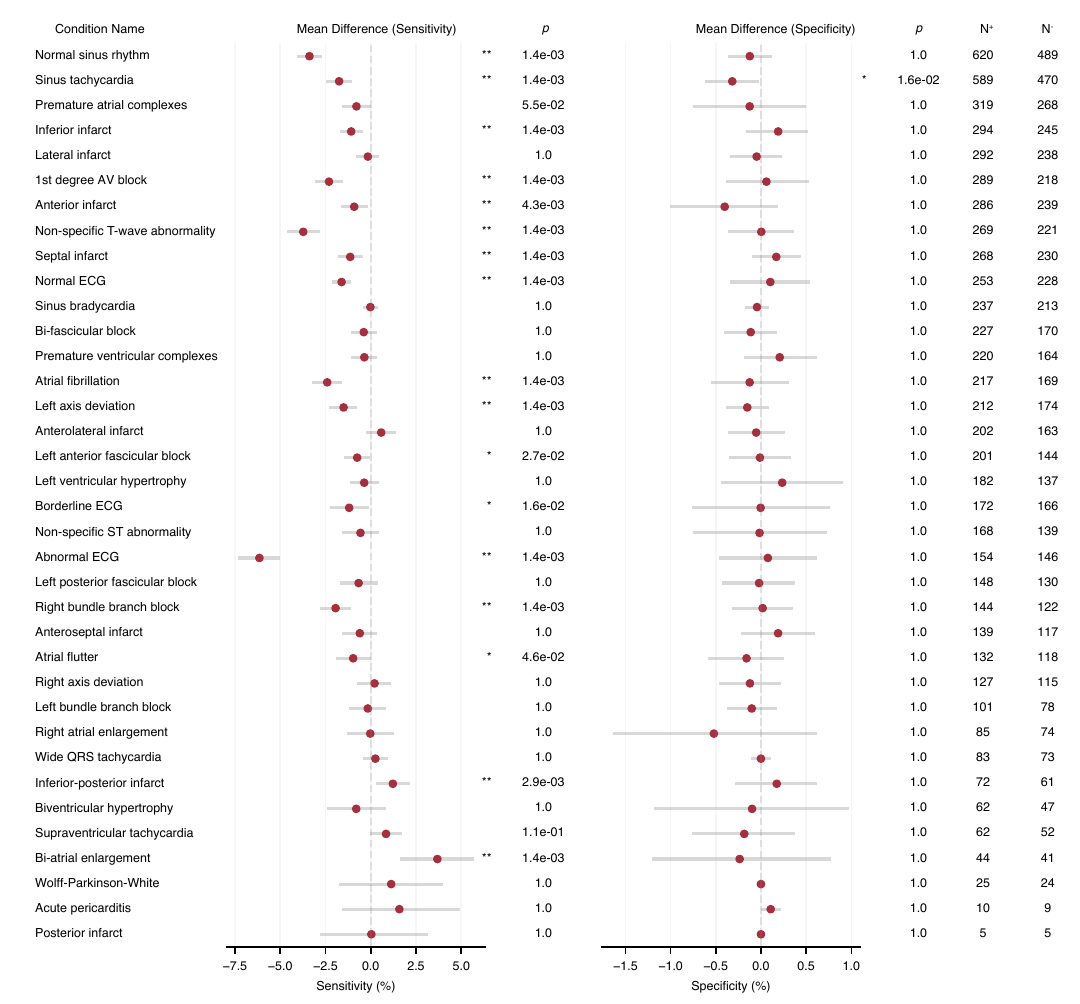}
    \vspace{1mm}
    \caption{\textbf{HEEDB data on diagnostic sensitivity and specificity changes on future \textit{de novo} cases for all available conditions.}
    Memorisation-induced changes in diagnostic sensitivity and specificity on future records of HEEDB patients returning with a \textit{de novo} condition, i.e.\@ one for which no positive cases were present in that patient's historical training record(s).
    Each row represents a single condition: the left-hand panel shows the mean difference in sensitivity between models trained vs.\ not trained on patients' historical data ($M=100$ models per group for each patient), the right-hand panel the corresponding difference in specificity.
    Point estimates to the left of zero indicate reduced performance on data-contributing patients' future records, those to the right an increase.
    $N^+$ and $N^-$ denote the number of positive and negative cases; cases comprised patients returning with a \textit{de novo} positive diagnosis, each matched to a randomly selected age- and sex-matched control where possible.
    Condition-specific decision thresholds maximised Youden's $J$ statistic on the test set.
    Statistical significance was determined using exact, two-sided permutation tests that reassign model training run indices to the corresponding patient subset masks ($B=100{,}000$ permutations per comparison); error bars denote simultaneous $95\%$ confidence intervals obtained by inverting the same test.
    $p$ values and confidence intervals were corrected across all $C=72$ shown comparisons using the Bonferroni method.
    * and ** denote $p\leq0.05$ and $p\leq0.01$, respectively; the resolution of the permutation test floors raw $p$ values at $2/(B+1)\approx2\times10^{-5}$, and hence corrected $p$ values at $1.4\times10^{-3}$.
    }
    \label{fig:edfig4}
\end{figure}

\begin{figure}
    \centering
    \includegraphics[width=\linewidth]{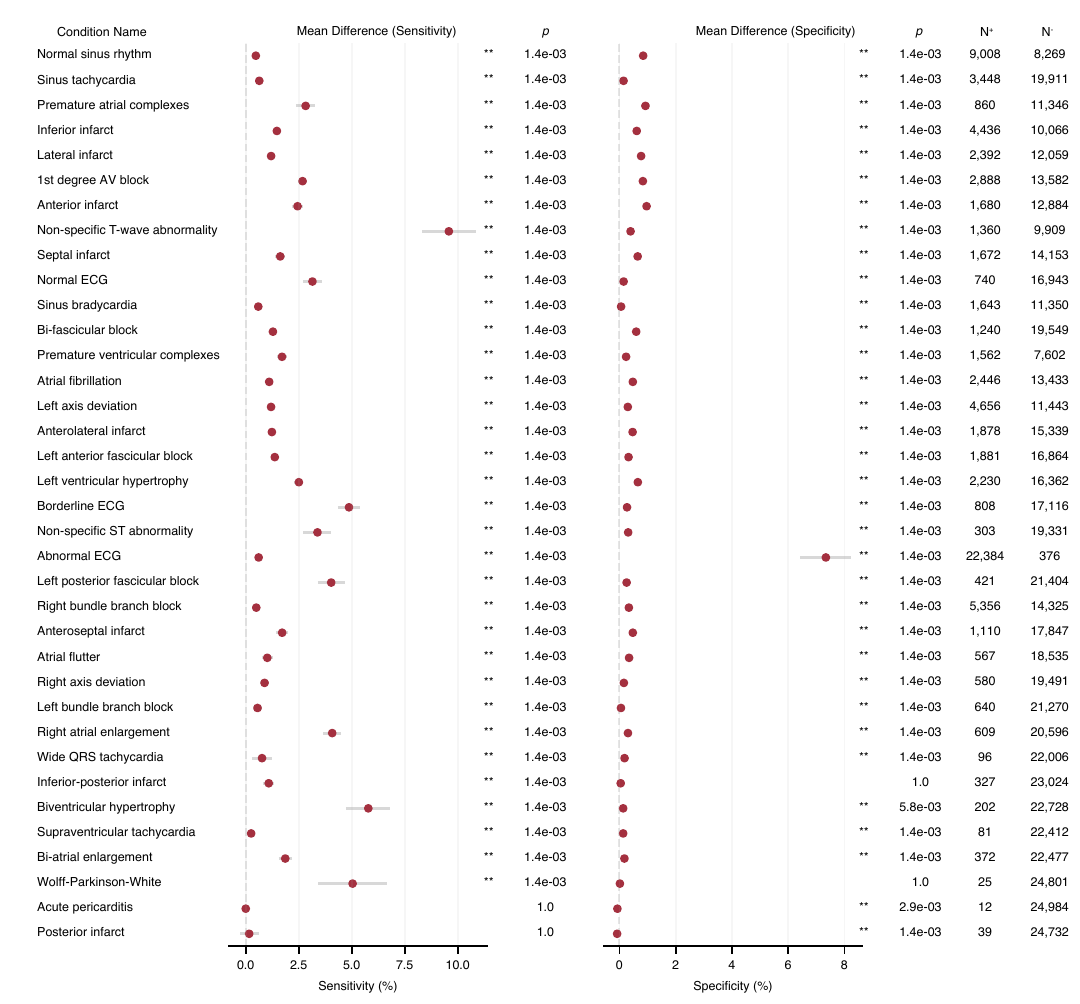}
    \vspace{1mm}
    \caption{\textbf{HEEDB data on diagnostic sensitivity and specificity changes on future records of patients returning with unchanged health states for all available conditions.}
    Memorisation-induced changes in diagnostic sensitivity and specificity on future records of HEEDB patients whose health state was unchanged relative to their historical training records; all $36$ available conditions are shown.
    Cases comprised positive future cases of a condition for which at least one positive case was already present in that patient's historical training records, and negative future cases of a condition for which only negative cases were present.
    Each row represents a single condition: the left-hand panel shows the mean difference in sensitivity between models trained vs.\ not trained on that patient's historical data ($M=100$ models per group for each patient), the right-hand panel the corresponding difference in specificity.
    Point estimates to the right of zero indicate increased performance on data-contributing patients’ future records, those to the left a reduction.
    $N^+$ and $N^-$ denote the number of positive and negative cases for each condition, respectively.
    Condition-specific decision thresholds maximised Youden's $J$ statistic on the test set.
    Statistical significance was determined using exact, two-sided permutation tests that reassign model training run indices to the corresponding patient subset masks ($B=100{,}000$ permutations per comparison); error bars denote simultaneous $95\%$ confidence intervals obtained by inverting the same test.
    $p$ values and confidence intervals were corrected across all $C=72$ shown comparisons using the Bonferroni method.
    * and ** denote $p\leq0.05$ and $p\leq0.01$, respectively; the resolution of the permutation test floors raw $p$ values at $2/(B+1)\approx2\times10^{-5}$, and hence corrected $p$ values at $1.4\times10^{-3}$.
    }
    \label{fig:edfig5}
\end{figure}

\begin{figure}
    \centering
    \includegraphics[width=\linewidth]{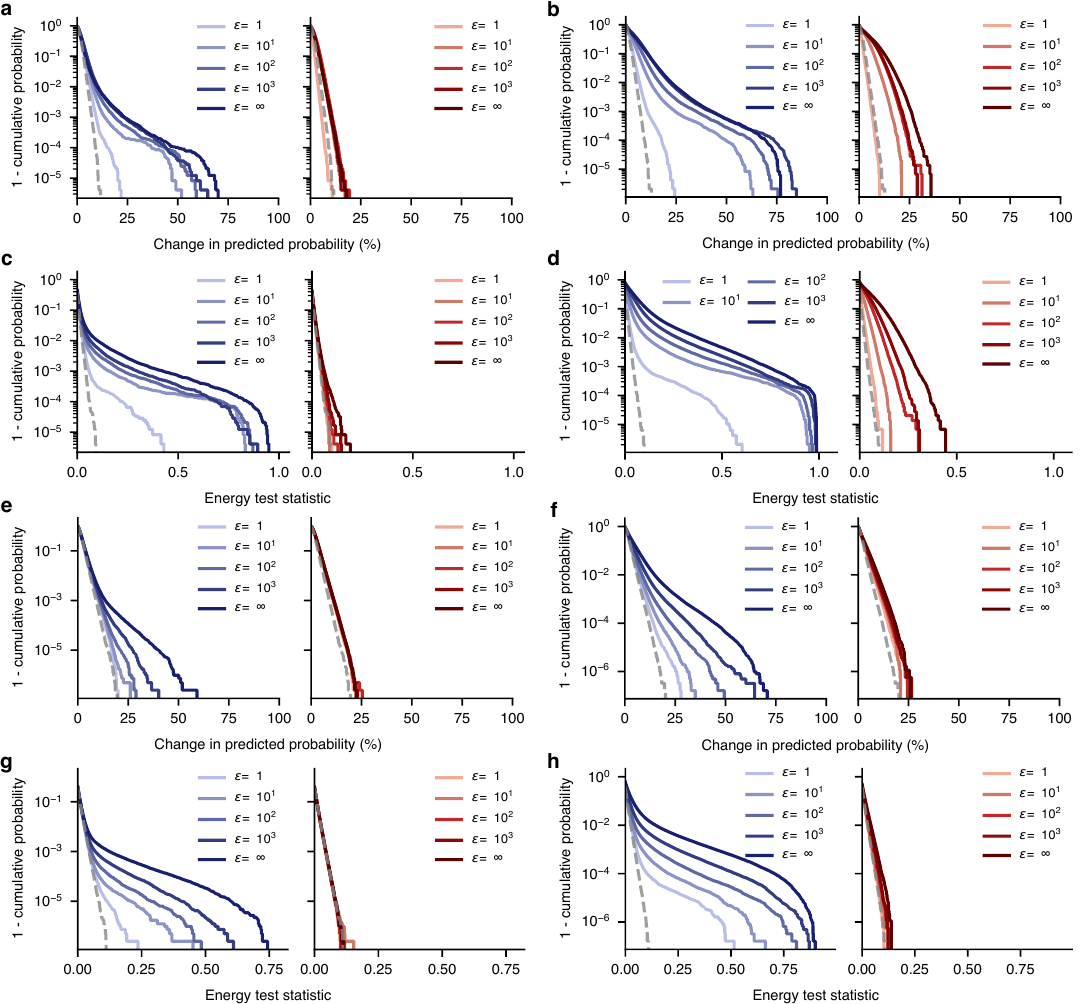}
    \caption{\textbf{Effect sizes and energy test statistics for models trained with different levels of record- and patient-level DP protection}
    \textbf{a-d}, results for MIMIC-ECG.
    \textbf{a-b}, ecCDF analysis of memorisation effect sizes for models trained with decreasing levels of ($\varepsilon$, $\delta$)-DP privacy protection (increasing $\varepsilon$ values).
    Effect size was measured as the absolute change in average predicted probability between models trained vs.\@ not-trained on a given patient's historical data (maximum across available classes).
    Panels show results for unseen future records (\textbf{a}) and for historical records used for model training (\textbf{b}).
    Each panel shows results for models protected by record-level DP (left, blue) and patient-level DP (right, red) separately; colour intensity increases with $\varepsilon$.
    \textbf{c-d}, as in panels (\textbf{a-b}), but for the energy test statistics from which the corresponding $p$ values were derived.
    \textbf{e-h}, as in panels (\textbf{a-d}) but for HEEDB.
    Dashed grey lines indicate the random-partitioning baseline.
    Effect sizes and test statistics were obtained from multivariate, nonparametric energy-based hypothesis tests comparing the predictions (predicted probabilities for all available classes) of models trained versus not trained on the respective patient's historical data ($M=100$ models per group for each patient).
    $\delta$ was kept constant at $1/D$ where $D$ is the dataset size.
    Patient-level DP protection was achieved by discarding all but the most recent historical training record per patient and then applying record-level DP accounting; the future record dataset was not modified.
    }
    \label{fig:edfig6}
\end{figure}

\clearpage

\section*{Supplementary Material}
\setcounter{figure}{0}
\setcounter{table}{0}
\beginsupplement
\listofsupptables

\begin{sidewaystable}[h]
    \centering
    \begin{tabular}{llccccccc}
         \textbf{Dataset} & \textbf{Model}  & \textbf{$\alpha$} & \textbf{$\lambda$} & \textbf{$\gamma$}& \textbf{Dropout} & \textbf{Batch size} & \textbf{Epochs} \\ \toprule
         MIMIC-ECG        & ViT-S-25        & $5\times10^{-3}$ & $0.5$      & $0.95$                    & $0.1$ & $256\ (1024)$ & $50$  \\
         MIMIC-ECG (DP)   & ViT-T-25        & $10^{-3}$        & $0$        & $0.99$                    & $0.0$ & $1024$        & $100$ &  \\
         MIMIC-CXR        & DenseNet-121    & $5\times10^{-4}$ & $10^{-2}$  & $0.9$                     & $0.0$ & $256$         & $60$  \\
         MIMIC-IV-ED      & ResNet          & $10^{-2}$        & $10^{0}$   & $0.99$                    & $0.25$ & $4096$       & $50$ \\
         HEEDB            & ViT-S-25        & $10^{-4}$        & $10^{-2}$  & $0.99$                    & $0.1$ & $256\ (512)$  & $20$   \\
         HEEDB (DP)       & ViT-T-25        & $3\times10^{-3}$ & $0$        & $0.99$                    & $0.0$ & $8192$        & $20$ \\
         \bottomrule
    \end{tabular}
    \caption[Model training hyperparameters.]{Model training hyperparameters. $\alpha$, $\lambda$ and $\gamma$ denote learning rate, weight decay and EMA decay rate, respectively. Where two batch sizes are given, the second is the effective batch size after gradient accumulation.}
    \label{tab:train_hypers}
\end{sidewaystable}





%\bibliography{bibliography}% common bib file

%TC:endignore

\end{document}